\documentclass[conference]{IEEEtran}
\IEEEoverridecommandlockouts
\usepackage{cite}
\usepackage{amsmath,amssymb,amsfonts}
\usepackage{algorithmic}
\usepackage{graphicx}
\usepackage{textcomp}
\usepackage{booktabs}
\usepackage{subcaption}
\usepackage{bm}
\usepackage{hyperref}
\usepackage[dvipsnames]{xcolor}
\usepackage{pifont}% http://ctan.org/pkg/pifont
\usepackage{enumitem}

\newcommand{\gcnhred}[1]{{\color{BrickRed} #1}}
\newcommand{\gcnhgreen}[1]{{\color{ForestGreen} #1}}
\newcommand{\cmark}{\ding{51}}%
\newcommand{\xmark}{\ding{55}}%

\def\BibTeX{{\rm B\kern-.05em{\sc i\kern-.025em b}\kern-.08em
    T\kern-.1667em\lower.7ex\hbox{E}\kern-.125emX}}
\begin{document}

% \title{GCNH: GCN for Heterophily}
\title{GCNH: A Simple Method For Representation Learning On Heterophilous Graphs} 

\author{\IEEEauthorblockN{Andrea Cavallo\IEEEauthorrefmark{1}\IEEEauthorrefmark{2},
Claas Grohnfeldt\IEEEauthorrefmark{2}, Michele Russo\IEEEauthorrefmark{2},
Giulio Lovisotto\IEEEauthorrefmark{2} and Luca Vassio\IEEEauthorrefmark{1}}
\IEEEauthorblockA{Politecnico di Torino\IEEEauthorrefmark{1}, Huawei Munich Research Center\IEEEauthorrefmark{2}\\
\IEEEauthorrefmark{1}\texttt{\{name\}.\{surname\}}@polito.it, \IEEEauthorrefmark{2}\texttt{\{name\}.\{surname\}}@huawei.com}}

% \author{\IEEEauthorblockN{Anonymous Authors}}
% \author{\IEEEauthorblockN{Andrea Cavallo}
% \IEEEauthorblockA{
% \textit{Politecnico di Torino}\\
% andrea.cavallo@polito.it}
% \and
% \IEEEauthorblockN{Claas Grohnfeldt}
% \IEEEauthorblockA{
% \textit{Huawei Munich Research Center}\\
% claas.grohnfeldt@huawei.com}
% \and
% \IEEEauthorblockN{Michele Russo}
% \IEEEauthorblockA{
% \textit{Huawei Munich Research Center}\\
% michele.russo1@huawei.com}
% \\
% \and
% \IEEEauthorblockN{Giulio Lovisotto}
% \IEEEauthorblockA{
% \textit{Huawei Munich Research Center}\\
% giulio.lovisotto@huawei.com}
% \and
% \IEEEauthorblockN{Luca Vassio}
% \IEEEauthorblockA{
% \textit{Politecnico di Torino}\\
% luca.vassio@polito.it}
% \and

\newcommand{\htwogcn}{H\textsubscript{2}GCN}

\maketitle

\begin{abstract}

Graph Neural Networks (GNNs) are well-suited for learning on homophilous graphs, i.e., graphs in which edges tend to connect nodes of the same type. Yet, achievement of consistent GNN performance on heterophilous graphs remains an open research problem. Recent works have proposed extensions to standard GNN architectures to improve performance on heterophilous graphs, trading off model simplicity for prediction accuracy.
However, these models fail to capture basic graph properties, such as neighborhood label distribution, which are fundamental for learning.

In this work, we propose GCN for Heterophily (GCNH), a simple yet effective GNN architecture applicable to both heterophilous and homophilous scenarios.
GCNH learns and combines \textit{separate} representations for a node and its neighbors, using one \textit{learned importance coefficient} per layer to balance the contributions of center nodes and neighborhoods. 
We conduct extensive experiments on eight real-world graphs and a set of synthetic graphs with varying degrees of heterophily to demonstrate how the design choices for GCNH lead to a sizable improvement over a vanilla GCN.
Moreover, GCNH outperforms state-of-the-art models of much higher complexity on four out of eight benchmarks, while producing comparable results on the remaining datasets.
Finally, we discuss and analyze the lower complexity of GCNH, which results in fewer trainable parameters and faster training times than other methods, and show how GCNH mitigates the oversmoothing problem.

\end{abstract}


\section{Introduction}\label{sec:introduction}

GNNs are core components of current state-of-the-art methods for learning and prediction on graph-structured data across domains and applications. Their capability to encode semantic and contextual information into node embeddings is known to be particularly effective on homophilous graphs, i.e., graphs in which nodes of the same type tend to be connected~\cite{kipf_semi_2017,velickovic_graph_2018}.
On the other hand, in the case of heterophilous graphs, in which neighboring nodes tend to be dissimilar in type, the achievement of competitive, or at least consistent, GNN prediction accuracy remains an open research goal~\cite{zhu_beyond_2020,jin_universal_2021}. 

Among other explanations for the inconsistent GNN performance in the heterophilous case, ~\cite{zheng_graph_2022} suggests that the message aggregation strategies of standard GNNs could lead to weakly representative node embeddings, due to a disproportionately high contribution of dissimilar neighbors.
That is why some works attempt to improve the generalization capability of GNNs to heterophilous networks by either modifying the graph structure or tailoring aggregation strategies and network architecture for such scenario~\cite{abu_mixhop_2019, zhu_beyond_2020, jin_universal_2021, wang_tree_2021,pei_geom-gcn_2019,suresh_breaking_2021,wei_node_2021,he_block_2022,yan_two_2021,bodnar_neural_2022,bo_beyond_2021,du_gbk_2022,yang_diverse_2021,chen_simple_2020,chien_adaptive_2021}.
Nevertheless, \cite{ma_is_2022} shows that a vanilla Graph Convolutional Network (GCN) can still \textit{outperform} more complex (heterophilous) models on some heterophilous graphs.
\cite{ma_is_2022} and~\cite{cavallo_2ncs_2022} suggest that this contradiction originates from the fact that the edge homophily ratio should not be considered a representative indicator of GCN performance, and instead recommend taking into account neighborhood structure and distribution of node labels.

Based on these observations, we propose a simple yet effective GNN architecture, namely GCN for Heterophily (GCNH), to improve node representation capabilities for applications on heterophilous graphs.
Differently from other GNNs, GCNH learns two different functions that encode a node and its neighbors \textit{separately}.
In addition, we allow the GCNH layer to flexibly assign different relevance to the information present in the neighborhood versus the information present in the center node.
We discuss and show how this design mitigates noisy neighborhood representations from negatively influencing the learned embeddings while allowing informative neighbors to strongly influence the final node embedding.
These extensions make GCNH more adaptive to heterophily than a vanilla GCN, while also improving prediction accuracy over more complex models designed for heterophily on common benchmarks.

We evaluate GCNH on the task of node classification using eight common real-world datasets and one set of synthetic graphs with varying degrees of heterophily.
We show that GCNH is able to learn meaningful representations of nodes independently of the homophily level of the graph, achieving new state-of-the-art performance on heterophilous graph datasets while producing results comparable to the state-of-the-art on homophilous benchmarks.

Our main contributions are summarized below.
\begin{itemize}
    \item We present GCNH, a simple yet effective GNN architecture that improves graph representation learning capabilities on heterophilous graphs while preserving the advantages of GCN.
    \item We present extensive experiments on real and synthetic datasets with varying degrees of heterophily for the node classification task. GCNH improves over the state-of-the-art on four (out of eight) real-world benchmarks while performing comparably to the state-of-the-art on all other datasets, including homophilous graphs.
    \item We showcase the lower complexity of GCNH compared to other state-of-the-art models, both in terms of the training time and the number of trainable parameters, and how it mitigates the oversmoothing problem.
\end{itemize}

\section{Related work}

The analysis and improvement of GNN performance on heterophilous graphs have received increasing attention in recent years~\cite{abu_mixhop_2019, zhu_beyond_2020, jin_universal_2021, wang_tree_2021,pei_geom-gcn_2019,suresh_breaking_2021,wei_node_2021,he_block_2022,yan_two_2021,bodnar_neural_2022,bo_beyond_2021,du_gbk_2022,yang_diverse_2021,chen_simple_2020,chien_adaptive_2021}.
GNNs generate node embeddings in two steps: message transformation, where node messages, i.e., features or embeddings from the previous layer, are transformed, and message aggregation, where the final node embeddings are generated by aggregating the messages of neighbors in the graph. One of the earliest and today most common GNN models is the GCN~\cite{kipf_semi_2017}, which defines message transformation as a learnable linear layer and message aggregation as the average of the messages from the neighbors and from the center node.

Several works have proposed new GNN architectures or graph transformations to make the message-passing framework suitable for applications on non-homophilous graphs. We compare a number of these approaches to our method in Section~\ref{sec:experiments}. Borrowing the categorization introduced in~\cite{zheng_graph_2022}, these approaches can be subdivided as follows.
\subsection{Non-Local Neighbor Extension}
These methods selectively extend the receptive field of GNNs to include potentially important nodes located outside of the local neighborhood. These approaches are based on the assumption that neighboring nodes may be dissimilar and that information about the target node, presumably carried by nodes with the same label, is available in nodes belonging to higher-order neighborhoods. In particular, methods such as~\cite{abu_mixhop_2019, zhu_beyond_2020, jin_universal_2021, wang_tree_2021} create latent representations for nodes using multiple neighborhoods at different hop distances and combine them into one overall embedding. Other approaches, such as~\cite{pei_geom-gcn_2019,suresh_breaking_2021,wei_node_2021,he_block_2022}, do not consider entire neighborhoods but single nodes as potential neighbors, regardless of their location in the graph, and aggregate messages from the nodes that are estimated to be more relevant. 
\subsection{GNN Architecture Adaptation}
Methods in this category modify the GNN architecture with the goal of improving representation learning capabilities on heterophilous graphs. A common approach is to estimate the relevance of a node to a given target node with respect to the prediction task at hand, and to assign individual weights to messages from neighbors based on their relevance. This approach is used in~\cite{yan_two_2021,bodnar_neural_2022,bo_beyond_2021,du_gbk_2022,yang_diverse_2021} among the others. Another approach is to learn separate embeddings for the neighborhood and the target node and merge them at a later stage. This mitigates the phenomenon of noisy embeddings, where information from potentially highly dissimilar neighbors is mixed with the features of the target node. Methods that follow this approach include~\cite{zhu_beyond_2020,suresh_breaking_2021}. 
In addition, \cite{zhu_beyond_2020,chen_simple_2020,chien_adaptive_2021} adopt the strategy of treating the information captured by different GNN layers separately, instead of first aggregating the information from all layers and then using the final node embeddings for prediction.

\section{Preliminaries}
\subsection{Notation And Problem Statement}
Let \(G = (V,E)\) be an unweighted and undirected graph, where \(V\) is the set of nodes and \(E\) is the set of edges. The connectivity information in the graph is represented by the adjacency matrix \(A\in \mathbb{R}^{n\times n}\), where \(n = |V|\) is the number of nodes and matrix elements \(A_{uv}\) are equal to 1 if nodes \(u\) and \(v\) are adjacent and 0 otherwise. Each node is associated with a feature vector \(x_u\) of size \(f\), and the complete set of features in the graph is denoted by \(X\in \mathbb{R}^{n\times f}\). The \textit{neighbors} of a node \(u\) are the set of nodes adjacent to it, denoted by \(\mathcal{N}_u\). Note that \(\mathcal{N}_u\) does not include node $u$.

The task addressed in this work is \textit{supervised node classification}. In this scenario, each node is associated with a label \(y_u\in C\) representing the class the node belongs to, where \(C\) is the set of labels. The task corresponds to learning a mapping \(\mathcal{F}: V \rightarrow C\) that uses the information of the graph \(G\), the features \(X\) and the labels to map nodes into their ground-truth class.
As is standard practice, we add a linear layer that maps the final node representations of the last network layer $H^{L} \in \mathbb{R}^{n\times e^L}$ (with $L$ total number of layers and $e^L$ size of the final node embeddings) to class probabilities with the addition of a softmax. For a node $u$:
\begin{equation}
    \label{eq:gcnh_class}
    \tilde{y}_{u} = \text{softmax}(h_u^{L} W_{cl}),
\end{equation}
where $W_{cl} \in  \mathbb{R}^{e^L \times |C|}$ is a learnable matrix. We omit the bias term from the equation for simplicity.
Note that $\tilde{y}_{u} \in \mathbb{R}^{|C|}$ is the probability distribution over classes for node $u$.
During inference, networks assign to the node the class with maximum probability:
% During inference, GCNH assigns to the node the class with maximum likelihood, i.e.:
~\begin{equation}
     \label{eq:gcnh_assigned_class}
     \hat{y}_{u} = \text{argmax}(\tilde{y}_{u}).
 \end{equation}
% We train the model to minimize the negative log-likelihood on the training data.
To train the model, we minimize the negative log-likelihood loss on the training data
\(\mathcal{L}(y_u,\hat{y_u})\),  where \(\hat{y_u} = \mathcal{F}(G, X, u)\) is the predicted label for node \(u\).
The loss function \(\mathcal{L}(\cdot,\cdot)\) describes the distance between the true label \(y_u\) and the predicted label. We focus on the transductive case: we work on single-graph and separate nodes into training, validation and test set.

\subsection{Homophily And Heterophily}
\textit{Graph homophily} is a social science-inspired property of graphs~\cite{mcpherson_birds_2001}, defined as the extent to which similar nodes in a graph are connected. Although other definitions of node similarity exist~\cite{yang_diverse_2021, ma_is_2022, cavallo_2ncs_2022}, this work focuses on label similarity, i.e., we define nodes as similar if they share the same label, and we measure graph homophily using the standard \textit{edge homophily ratio}, denoted by $h$:
\begin{equation}
    h = \frac{|\{(u,v):(u,v)\in E \land y_u=y_v\}|}{|E|},
    \label{eq:edge_hom_rat}
\end{equation}
which quantifies the fraction of edges in a graph that connect nodes with the same label.
Graphs with low values of \textit{h} are called \textit{heterophilous} or \textit{disassortative}. As discussed in Section~\ref{sec:introduction}, GNNs perform inconsistently on this category of graphs -- we analyze this point in depth in Section~\ref{sec:experiments}.

\begin{figure*}[t]
\centering
\includegraphics[width=0.8\linewidth]{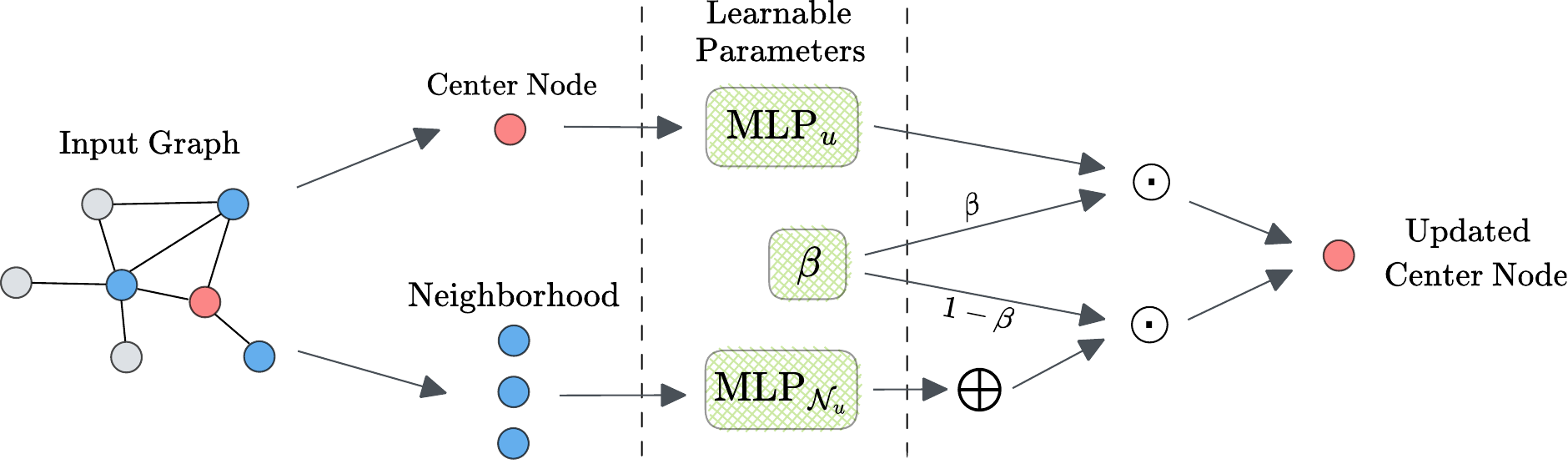}
\caption{Architecture of the GCNH layer. To create an updated representation for the center node (red), the GCNH layer explicitly separates the node from its 1-hop neighborhood (blue). Two different MLPs encode center node and neighborhood separately ($\text{\textsc{mlp}}_u$ and $\text{\textsc{mlp}}_{\mathcal{N}_u}$,  respectively). The encoded neighbors are then aggregated with a permutation invariant function $\bigoplus$. Finally, the aggregated neighborhood and the encoded center node are combined together with a \textit{learnable} weighting factor $\beta$ which regulates the contributions of the center node versus its neighborhood to produce the final output embedding for the node.}
\label{fig:gcnh_architecture2}
\end{figure*}

\section{Method}\label{sec:method}
In this section, we describe the architecture of the model proposed in this paper, namely GCNH. Figure~\ref{fig:gcnh_architecture2} illustrates the structure of the GCNH layer.

\subsection{GCNH Layer Formulation}
A GCNH network is composed of one or more, $L$, GCNH layers.
The $\ell^{\text{th}}$ layer receives in input the adjacency matrix and the node representations computed at the previous layer  $H^{\ell-1}$ and produces updated node representations $H^{\ell}$. We set $H^0=X$.

Within a layer, node representations are transformed through two separate 1-layer MLPs, $\text{\textsc{mlp}}_u$ and $\text{\textsc{mlp}}_{\mathcal{N}_u}$, 
resulting in latent representations $z_u$ and $z_{\mathcal{N}_u}$ of the target node and its neighborhood, respectively.
For a node $u$, at layer $\ell$, we formally describe this step as follows. First, an intermediate representation is computed for the node $u$:
\begin{equation}
    \label{eq:gcnh_feat_trans1b}
    z_u^{\ell} = \text{\textsc{mlp}}_u(h_u^{\ell-1}) = \sigma(h_u^{\ell-1}W_1).
\end{equation}
Secondly, all the representations of $u$'s neighbors ($v \in \mathcal{N}_u$) are  updated based on their current representations $h_v$ and aggregated together to obtain a neighborhood representation:
\begin{equation}
    \label{eq:gcnh_feat_trans2b}
    z_{\mathcal{N}_u}^{\ell} = \bigoplus_{v\in \mathcal{N}_u}(\text{\textsc{mlp}}_{\mathcal{N}_u}(h_v^{\ell-1})) = 	\bigoplus_{v\in \mathcal{N}_u}(\sigma(h_v^{\ell-1}W_2)).
\end{equation}
In Equations~\ref{eq:gcnh_feat_trans1b} and~\ref{eq:gcnh_feat_trans2b}, $W_1, W_2 \in \mathbb{R}^{e^{\ell-1} \times e^{\ell}}$ are learnable matrices, $\sigma(\cdot)$ is a generic activation function, $e^{\ell}$ is the size of the embeddings created at the $\ell$-th layer and $\bigoplus$ is a permutation invariant aggregation function over the nodes $v \in \mathcal{N}_u$. We omit the bias terms of the MLPs in the equations for clarity.
The final output embedding for node $u$ is obtained as a linear combination of $z_u^{\ell}$ and $z_{\mathcal{N}_u}^{\ell}$, parametrized by a learnable scalar value $\beta$:
\begin{equation}
    \label{eq:gcnh_aggr}
    h_u^{\ell} = (1-\beta) z_{\mathcal{N}_u}^{\ell} + \beta z_u^{\ell}
\end{equation}
where $\beta$ is normalized between 0 and 1 with a sigmoid$(\cdot)$ function.
Note that the MLPs, $\bigoplus$ and $\beta$ are different for each layer $\ell$. We omit superscripts in the equations for clarity.

\begin{table*}[t]
    \footnotesize
    \centering
    \addtolength{\leftskip} {-2cm}
    \addtolength{\rightskip}{-2cm}
    %\setcellgapes{3pt}
    %\makegapedcells
    \begin{tabular}{ c c c c c c c c c }
    \toprule
    \textbf{Benchmark} & \textbf{Cornell} & \textbf{Texas} & \textbf{Wisconsin} & \textbf{Film} & \textbf{Chameleon} & \textbf{Squirrel} & \textbf{Cora} & \textbf{Citeseer} \\
    \(h\) & 0.30 & 0.11 & 0.21 & 0.22 & 0.23 & 0.22 & 0.81 & 0.74 \\
    \midrule
    \textbf{MLP} & 81.89$\pm$6.40 & 80.81$\pm$4.75 & 85.29$\pm$3.31 & 36.53$\pm$0.70 & 46.21$\pm$2.99 & 28.77$\pm$1.56 & 75.69$\pm$2.00 & 74.02$\pm$1.90 \\
    \textbf{GCN} & 60.54$\pm$5.30 & 55.14$\pm$5.16 & 51.76$\pm$3.06 & 27.32$\pm$1.10 & 64.82$\pm$2.24 & 53.43$\pm$2.01 & 86.98$\pm$1.27 & 76.50$\pm$1.36 \\
    \textbf{GAT} & 61.89$\pm$5.05 & 52.16$\pm$6.63 & 49.41$\pm$4.09 & 27.44$\pm$0.89 & 60.26$\pm$2.50 & 40.72$\pm$1.55 & 87.30$\pm$1.10 & 76.55$\pm$1.23 \\
    \textbf{Geom-GCN} & 60.54$\pm$3.67 & 66.76$\pm$2.72 & 64.51$\pm$3.66 & 31.59$\pm$1.15 & 60.00$\pm$2.81 & 43.80$\pm$1.48 & 85.35$\pm$1.57 & \textbf{\textcolor{red}{78.02$\pm$1.15}} \\
    \textbf{\htwogcn{}} & 82.70$\pm$5.28 & 84.86$\pm$7.23 & \textbf{\textcolor{violet}{87.65$\pm$4.98}} & 35.70$\pm$1.00 & 60.11$\pm$2.15 & 36.48$\pm$1.86 & \textbf{\textcolor{violet}{87.87$\pm$1.20}} & 77.11$\pm$1.57 \\
    \textbf{GPRGNN} & 80.27$\pm$8.11 & 78.38$\pm$4.36 & 82.94$\pm$4.21 & 34.63$\pm$1.22 & 46.58$\pm$1.71 & 31.61$\pm$1.24 & \textbf{\textcolor{blue}{87.95$\pm$1.18}} & \textbf{\textcolor{violet}{77.13$\pm$1.67}} \\
    \textbf{GGCN} & \textbf{\textcolor{blue}{85.68$\pm$6.63}} & \textbf{\textcolor{violet}{84.86$\pm$4.55}} & 86.86$\pm$3.29 & \textbf{\textcolor{blue}{37.54$\pm$1.56}} & \textbf{\textcolor{blue}{71.14$\pm$1.84}} & \textbf{\textcolor{violet}{55.17$\pm$1.58}} & \textbf{\textcolor{red}{87.95$\pm$1.05}} & \textbf{\textcolor{blue}{77.14$\pm$1.45}} \\
    \textbf{O(d)-SD} & \textbf{\textcolor{violet}{84.86$\pm$4.71}} & \textbf{\textcolor{blue}{85.95$\pm$5.51}} & \textbf{\textcolor{red}{89.41$\pm$4.74}} & \textbf{\textcolor{red}{37.81$\pm$1.15}} & \textbf{\textcolor{violet}{68.04$\pm$1.58}} & \textbf{\textcolor{blue}{56.34$\pm$1.32}} & 86.90$\pm$1.13 & 76.70$\pm$1.57 \\
    \midrule
    \textbf{GCNH} & \textbf{\textcolor{red}{86.49$\pm$6.98}} & \textbf{\textcolor{red}{87.84$\pm$3.87}} & \textbf{\textcolor{blue}{87.65$\pm$3.59}} & \textbf{\textcolor{violet}{36.89$\pm$1.50}} & \textbf{\textcolor{red}{71.56$\pm$1.86}} & \textbf{\textcolor{red}{61.85$\pm$1.54}} & 86.88$\pm$1.04 & 75.81$\pm$1.14 \\
    % \textbf{GCNH-mean} & \textbf{\textcolor{blue}{86.49$\pm$4.41}} & \textbf{\textcolor{blue}{88.92$\pm$3.24}} & \textbf{\textcolor{red}{89.80$\pm$5.29}} & 36.83$\pm$1.44 & 55.64$\pm$1.95 & 39.24$\pm$1.02 & 85.75$\pm$1.57 & 76.00$\pm$0.99 \\
    % \textbf{GCNH-maxpool} & \textbf{\textcolor{red}{87.30$\pm$5.98}} & \textbf{\textcolor{red}{89.73$\pm$4.38}} & \textbf{\textcolor{violet}{88.63$\pm$4.01}} & 36.28$\pm$1.26 & \textbf{\textcolor{red}{72.50$\pm$1.60}} & \textbf{\textcolor{red}{64.56$\pm$1.39}} & 86.68$\pm$1.10 & 75.96$\pm$1.33 \\
    \bottomrule

    \end{tabular}
    \caption{Mean classification accuracy and standard deviation for GCNH on real-world datasets, on the 10 splits taken from~\cite{pei_geom-gcn_2019}. Best results are in \textbf{\textcolor{red}{red}}, second best results in \textbf{\textcolor{blue}{blue}} and third best in \textbf{\textcolor{violet}{violet}}. The results for the baselines are taken from~\cite{yan_two_2021} and~\cite{bodnar_neural_2022}. We use sum as the aggregation function in GCNH; other parameters of GCNH are selected from the best-performing configuration (see Table~\ref{tab:gcnh_hyp} in the Appendix for details on the hyperparameters). }
    \label{tab:main_results}
\end{table*}

\subsection{GCNH Design Choices}
\label{sec:gcnh-design}

%We describe the rationale behind GCNH in the following.
GCNH introduces two main design choices which differentiate it from a standard GCN: (i) the separate encoding of the target node and its neighbors (Equations~\ref{eq:gcnh_feat_trans1b}, ~\ref{eq:gcnh_feat_trans2b}) and (ii) the explicit parameterization of the contributions of neighborhood and center node with $\beta$ (Equation~\ref{eq:gcnh_aggr}). 

The positive impact of separately processing the node and its neighborhood has been previously outlined in~\cite{zhu_beyond_2020,suresh_breaking_2021}.
Intuitively, in homophilous settings, where neighbors are similar, aggregation from the neighborhood brings useful information. On the other hand, in heterophilous graphs, dissimilar neighbors might bring detrimental information that a GNN cannot easily ignore.
Compared to~\cite{zhu_beyond_2020}, GCNH takes the separation principle further, while minimizing the amount of complexity that this choice adds to the network architecture.
Specifically, three choices distinguish GCNH from the model proposed in~\cite{zhu_beyond_2020}: (i) we only use 1-hop neighborhoods, (ii) we learn separate MLPs for center node and neighborhood and (iii) we combine separate embeddings using a learned linear combination instead of concatenation.
Note that (i) leads to faster computation while retaining most of the information useful for node representation, as shown in Section~\ref{sec:experiments}, (ii) improves flexibility and (iii) leads to models with fewer parameters. (i) and (iii) lead to lower time complexity compared to~\cite{zhu_beyond_2020} (see Section~\ref{sec:complexity} for a time complexity analysis).

Combined with the embedding separation, the explicit modeling of the importance of the neighborhood informativeness with the coefficient $\beta$ in GCNH allows to adaptively determine the impact of the neighborhood on the final node embeddings based on how informative the neighbors are.
Note that informative neighbors might also exist in heterophilous graphs~\cite{ma_is_2022, cavallo_2ncs_2022}; therefore, the contribution of neighbors is not necessarily related to the homophily level of the network.
Intuitively, modeling and learning $\beta$ explicitly provides a helpful inductive bias that allows the network to directly prefer the center node versus neighborhood information. Section~\ref{sec:experiments} shows how models that are equipped with similar, or better, flexibility (such as GAT~\cite{velickovic_graph_2018}) perform poorly in comparison~\cite{lim_large_2021}.

\subsection{Time Complexity}\label{sec:complexity}
We compute the time complexity of a generic GCNH layer $\ell$ by analyzing the individual processing steps of the model.

First, the node representations $H^{\ell-1}$ are transformed, separately for the center node and the neighbors, as described in Equations~\eqref{eq:gcnh_feat_trans1b} and \eqref{eq:gcnh_feat_trans2b}. This step has a time complexity of $\mathcal{O}\left(ne^{\ell-1}e^{\ell}\right)$. Subsequently, the neighbors' representations are aggregated and merged with the center-node embedding according to Equation~\eqref{eq:gcnh_aggr}. All the neighbor aggregation functions tested (see Section \ref{sec:agg_func}) have a time complexity of $\mathcal{O}\left(|E|e^{\ell}\right)$, whereas the weighted sum with the self-node representation has a time complexity of $\mathcal{O}\left(ne^{\ell}\right)$. This last term is dominated by the complexity of the transformation. In total, the overall time complexity of the GCNH layer is $\mathcal{O}\left(ne^{\ell-1}e^{\ell} + |E|e^{\ell}\right)$. The linear dependency of the complexity of the GCNH layer on the number of nodes $n$ makes it easier to scale on large graphs compared to attention-based models (e.g., GGCN~\cite{yan_two_2021}), whose complexity depends quadratically on $n$.

\section{Experiments}\label{sec:experiments}
 
In this section, we analyze the learning capability of GCNH on the task of supervised node classification, commonly used for models dealing with heterophilous graphs. We perform the evaluation both on synthetic datasets with different levels of homophily ratio and on the real-world graphs widely used in related works. Further information about the datasets is reported in Appendix \ref{app:datasets}.

\subsection{Baselines}
The baselines used for comparison belong to two main categories. The first category includes well-known methods:
\begin{itemize}
    \item \textbf{MLP}: Multi-Layer Perceptron.
    \item \textbf{GCN} \cite{kipf_semi_2017}: Graph Convolutional Network.
    \item \textbf{GAT} \cite{velickovic_graph_2018}: Graph Attention Network.
\end{itemize}
Methods in the second category, instead, are specifically designed for heterophilous graphs. In particular:
\begin{itemize}
    \item \textbf{Geom-GCN} \cite{pei_geom-gcn_2019} maps nodes to a latent space and defines a new graph based on embedding similarity.
    \item \textbf{H\textsubscript{2}GCN} \cite{zhu_beyond_2020} introduces three specific designs to boost performance on heterophilous graphs: ego and neighbor-embedding separation, higher-order neighborhoods and combination of intermediate representations.
    \item \textbf{GPRGNN} \cite{chien_adaptive_2021} uses PageRank to determine relations between nodes.
    \item \textbf{GGCN} \cite{yan_two_2021} applies two designs to extend GNNs: degree correction and signed messages.
    \item \textbf{O(d)-SD} \cite{bodnar_neural_2022} is based on sheaf diffusion.
\end{itemize}

\subsection{Experimental Setting}
We evaluate model performance in terms of classification accuracy, i.e., the percentage of nodes in the test set that are assigned to the correct class.
We test different values for hyperparameters and we select the best ones. We report further information about the grids for the hyperparameters in Appendix~\ref{app:hyp}. The aggregation function $\bigoplus$ in Equation~\eqref{eq:gcnh_feat_trans2b} is an element-wise sum unless differently specified (see Section~\ref{sec:agg_func} for a comparison of aggregation functions).
We compute classification accuracies on 10 different train/validation/test splits provided by \cite{pei_geom-gcn_2019} for each dataset, and we report mean and standard deviation. The sizes of the splits are 48\%/32\%/20\%. For each dataset, we train the models on the training sets and the model performing best on average across the validation sets is used on the test sets, on which we compute the accuracy values.

% We run experiments on an NVIDIA Tesla V100 PCIE with 16 GB, except for the experiments in Table \ref{tab:training times} that we run on an NVIDIA Tesla T4. The code is implemented in Python and deep learning models are defined and trained using PyTorch. Our code is available at \url{https://github.com/SmartData-Polito/GCNH}.
We run experiments on an NVIDIA Tesla V100 PCIE with 16 GB, except for the experiments in Table \ref{tab:training times} that we run on an NVIDIA Tesla T4. The code is implemented in Python and PyTorch is used for deep learning models. Our code is available at \url{https://github.com/SmartData-Polito/GCNH}.

\subsection{Results On Real-World Datasets}
Table~\ref{tab:main_results} reports classification accuracies for GCNH and baselines on real-world datasets. 
GCNH achieves state-of-the-art performance on four out of the eight datasets used; it ranks second and third on Wisconsin and Film and performs slightly worse on the two homophilous graphs Cora and Citeseer.
The results on all the heterophilous graphs prove the effectiveness of the design choices of GCNH. 
On Cornell, Texas, Wisconsin and Film, GCNH avoids the detrimental influence of neighbors on the final embeddings observed, for example, in GCN and GAT, with respect to which improvements are large (up to 35\%).
On Chameleon and Squirrel, GCNH manages to encode the useful neighborhood information present in the graph thanks to its separate processing of node and 1-hop neighbors. Indeed, on these two graphs, GCN performs quite well (it ranks fourth on both), meaning that information contained in the 1-hop neighborhood is useful for node classification, whereas more complex models (e.g. Geom-GCN, \htwogcn{} and GPRGNN) fail to capture this property and perform worse (from 5 to 20\% accuracy drop with respect to GCN). 
On the homophilous graphs Cora and Citeseer, GCNH performs comparably to the other models, although slightly worse. 
These results show that GCNH flexibly adapts to various homophily settings, achieving consistent performance.

\begin{figure}[t]
\centering
\includegraphics[width=\linewidth]{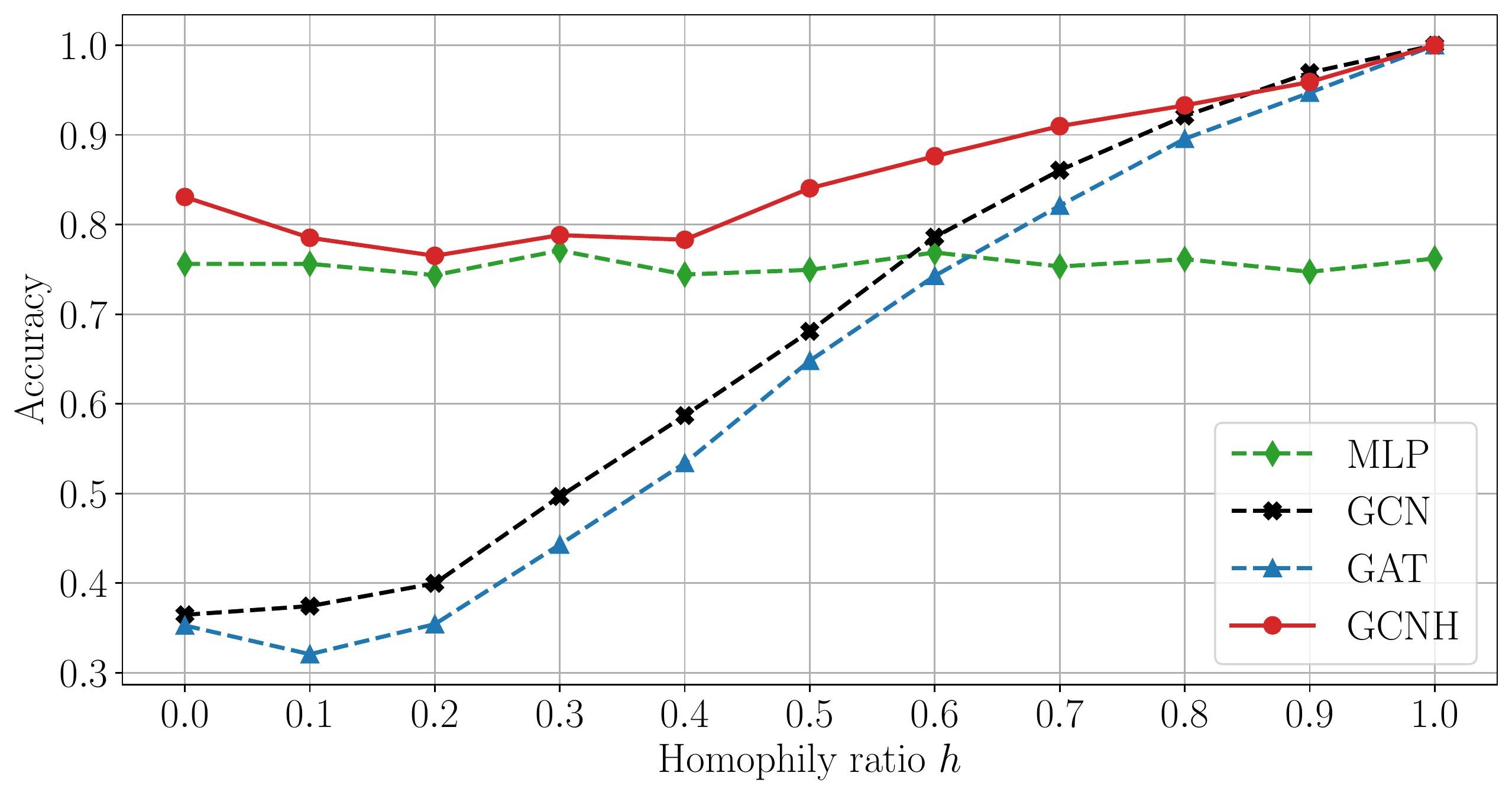}
\caption{Comparison of the classification accuracy achieved by a set of models on the syn-cora synthetic datasets. Each point shows the average accuracy on three datasets generated with a specific homophily ratio.}
\label{fig:gcnh_syn}
\end{figure}

\subsection{Results On Synthetic Datasets}

To better understand how GCNH deals with different levels of homophily, we evaluate it on synthetic graphs covering a wide range of values of $h$, while keeping node features and other graph properties unchanged. 
We provide additional details about the dataset used in Appendix~\ref{app:dataset:syn}.

Figure~\ref{fig:gcnh_syn} shows the accuracy of GCNH and three other simple baselines: MLP, whose performance is not affected by the value of $h$ as it is a graph-unaware method, and GCN and GAT, which achieve significantly worse results on heterophilous settings.
The hyperparameter configurations for these baselines are reported in Appendix~\ref{app:baselines_syn_hyp}. 
GCNH's performance is significantly less dependent on the homophily level of the graph than the performances of GAT and GCN. 
On heterophilous graphs, GCNH improves by almost 50\% over GCN and GAT and by $\sim$2/8\% over MLP, meaning that the separate encoding of the neighbors has a positive impact. 
On homophilous graphs, GCNH performs comparably to GCN and GAT, achieving perfect accuracy on perfectly homophilous graphs ($h=1$).

\begin{table*}[t]
    \footnotesize
    \centering
    \addtolength{\leftskip} {-2cm}
    \addtolength{\rightskip}{-2cm}
    \setlength{\tabcolsep}{0.44em}
    %\setcellgapes{3pt}
    %\makegapedcells
    \begin{tabular}{ c c c | c c c c c c c c }
    \toprule
    & Separate MLPs & Learned $\beta$  &\textbf{Cornell} & \textbf{Texas} & \textbf{Wisconsin} & \textbf{Film} & \textbf{Chameleon} & \textbf{Squirrel} & \textbf{Cora} & \textbf{Citeseer} \\
    % \(h\) & 0.30 & 0.11 & 0.21 & 0.22 & 0.23 & 0.22 & 0.81 & 0.74 \\
    \midrule
\textbf{GCN} & \xmark & \xmark & 60.54 & 55.14 & 51.76 & 27.32 & 64.82 & 53.43 & \textbf{86.98} & \textbf{76.50} \\\midrule 
\textbf{GCNH} & \cmark & \xmark & 83.78(\gcnhgreen{$+$23.2}) & 86.49(\gcnhgreen{$+$31.3}) & 85.49(\gcnhgreen{$+$33.7}) & 36.01(\gcnhgreen{$+$8.7}) & 70.22(\gcnhgreen{$+$5.4}) & 59.74(\gcnhgreen{$+$6.3}) & 86.90(\gcnhred{$-$0.1}) & 75.65(\gcnhred{$-$0.8}) \\ 
% GCNH (pointwise $\beta$) & 85.41(\gcnhgreen{$+$24.9}) & 85.95(\gcnhgreen{$+$30.8}) & \textbf{87.65}(\gcnhgreen{$+$35.9}) & 36.21(\gcnhgreen{$+$8.9}) & 70.31(\gcnhgreen{$+$5.5}) & 60.10(\gcnhgreen{$+$6.7}) & 86.46(\gcnhred{$-$0.5}) & 75.09(\gcnhred{$-$1.4}) \\ 
\textbf{GCNH} & \cmark & \cmark & \textbf{86.49}(\gcnhgreen{$+$25.9}) & \textbf{87.84}(\gcnhgreen{$+$32.7}) & \textbf{87.65}(\gcnhgreen{$+$35.9}) & \textbf{36.89}(\gcnhgreen{$+$9.6}) & \textbf{71.56}(\gcnhgreen{$+$6.7}) & \textbf{61.85}(\gcnhgreen{$+$8.4}) & 86.88(\gcnhred{$-$0.1}) & 75.81(\gcnhred{$-$0.7}) \\  
    \bottomrule
    \end{tabular}
    \caption{Mean classification accuracy of GCN and ablated GCNH. In parentheses, we report the performance \gcnhgreen{improvement} or \gcnhred{degradation} from the GCN baseline. ``Separate MLPs'' refers to whether we learn separate linear layers for node and neighborhoods ($W_1, W_2$ in Equation~\ref{eq:gcnh_feat_trans1b} and~\ref{eq:gcnh_feat_trans2b}, respectively). ``Learned $\beta$'' refers to whether we learn $\beta$ as a parameter or not; if not, it is fixed at $\beta$=0.5. Best results are in \textbf{bold}.}
    \label{tab:ablation_exp}
    
\end{table*}

% \begin{table*}[t]
%     \footnotesize
%     \centering
%     \addtolength{\leftskip} {-2cm}
%     \addtolength{\rightskip}{-2cm}
%     %\setcellgapes{3pt}
%     %\makegapedcells
%     \begin{tabular}{ c c c c c c c c c }
%     \toprule
%     \textbf{Benchmark} & \textbf{Cornell} & \textbf{Texas} & \textbf{Wisconsin} & \textbf{Film} & \textbf{Chameleon} & \textbf{Squirrel} & \textbf{Cora} & \textbf{Citeseer} \\
%     \(h\) & 0.30 & 0.11 & 0.21 & 0.22 & 0.23 & 0.22 & 0.81 & 0.74 \\
%     \midrule
%     \textbf{GCN} & 60.54$\pm$5.30 & 55.14$\pm$5.16 & 51.76$\pm$3.06 & 27.32$\pm$1.10 & 64.82$\pm$2.24 & 53.43$\pm$2.01 & \textbf{86.98$\pm$1.27} & \textbf{76.50$\pm$1.36} \\
%     \textbf{GCNH} ($\beta$=0.5) & 83.78$\pm$7.21 & 	86.49$\pm$5.98 & 85.49$\pm$5.56  & 36.01$\pm$1.47 & 70.22$\pm$1.62 & 59.74$\pm$1.83 & 86.90$\pm$1.35 & 75.65$\pm$1.04 \\
%     \textbf{GCNH} (pointwise $\beta$) & 85.41$\pm$5.87 & 85.95$\pm$5.22 & \textbf{87.65$\pm$3.07}  & 36.21$\pm$1.39 & 70.31$\pm$2.01 & 60.10$\pm$2.07 & 86.46$\pm$0.80 & 75.09$\pm$2.15 \\
%     \textbf{GCNH} & \textbf{86.49$\pm$ 6.98} & \textbf{87.84$\pm$3.87} & \textbf{87.65}$\pm$\textbf{3.59} & \textbf{36.89$\pm$1.50} & \textbf{71.56$\pm$1.86} & \textbf{61.85$\pm$1.54} & 86.88$\pm$1.04 & 75.81$\pm$1.14 \\
%     \bottomrule
%     \end{tabular}
%     \caption{\textbf{(TODO: updated caption and reference in paper body)} Mean classification accuracy and standard deviation for GCN and GCNH with fixed $\beta$ or with different aggregation functions. Best results are in \textbf{bold}.}
%     \label{tab:ablation_exp}
    
% \end{table*}
\begin{table*}[t]
    \footnotesize
    \centering
    \addtolength{\leftskip} {-2cm}
    \addtolength{\rightskip}{-2cm}
    %\setcellgapes{3pt}
    %\makegapedcells
    \begin{tabular}{ c c c c c c c c c }
    \toprule
    \textbf{Benchmark} & \textbf{Cornell} & \textbf{Texas} & \textbf{Wisconsin} & \textbf{Film} & \textbf{Chameleon} & \textbf{Squirrel} & \textbf{Cora} & \textbf{Citeseer} \\
    % \(h\) & 0.30 & 0.11 & 0.21 & 0.22 & 0.23 & 0.22 & 0.81 & 0.74 \\
    \midrule
    \textbf{GCNH} (sum) & 86.49$\pm$6.98 & 87.84$\pm$3.87 & 87.65$\pm$3.59 & \textbf{36.89$\pm$1.50} & 71.56$\pm$1.86 & 61.85$\pm$1.54 & \textbf{86.88$\pm$1.04} & 75.81$\pm$1.14 \\
    \textbf{GCNH} (mean) & \textbf{86.49$\pm$4.41} & \textbf{88.92$\pm$3.24} & \textbf{89.80$\pm$5.29} & 36.83$\pm$1.44 & 55.64$\pm$1.95 & 39.24$\pm$1.02 & 85.75$\pm$1.57 & \textbf{76.00$\pm$0.99} \\
    % \textbf{GCNH} (max) & \textbf{87.30$\pm$5.98} & \textbf{89.73$\pm$4.38} & 88.63$\pm$4.01 & 36.74$\pm$1.17 & \textbf{72.50$\pm$1.60} & \textbf{64.56$\pm$1.39} & 86.68$\pm$1.10 & 75.96$\pm$1.33 \\
    \textbf{GCNH} (max) & 85.41$\pm$8.18 & 88.65$\pm$4.56 & 88.04$\pm$3.26 & 	36.07$\pm$1.26 & \textbf{71.64$\pm$1.85} & \textbf{63.35$\pm$1.99} & 86.68$\pm$1.10 & 75.96$\pm$1.33 \\
    \bottomrule

    \end{tabular}
    \caption{Mean classification accuracy and standard deviation for GCNH with different aggregation functions in Equation~\ref{eq:gcnh_feat_trans2b}. Note that results for ``max'' are obtained full-batch training only (see Table~\ref{tab:gcnh_hyp} and discussion in Section~\ref{sec:training_times}), as it allows for more efficient implementation. Best results are in \textbf{bold}.}
    \label{tab:aggfunc_exp}
\end{table*}

\begin{figure}[t]
\centering
\includegraphics[width=\linewidth]{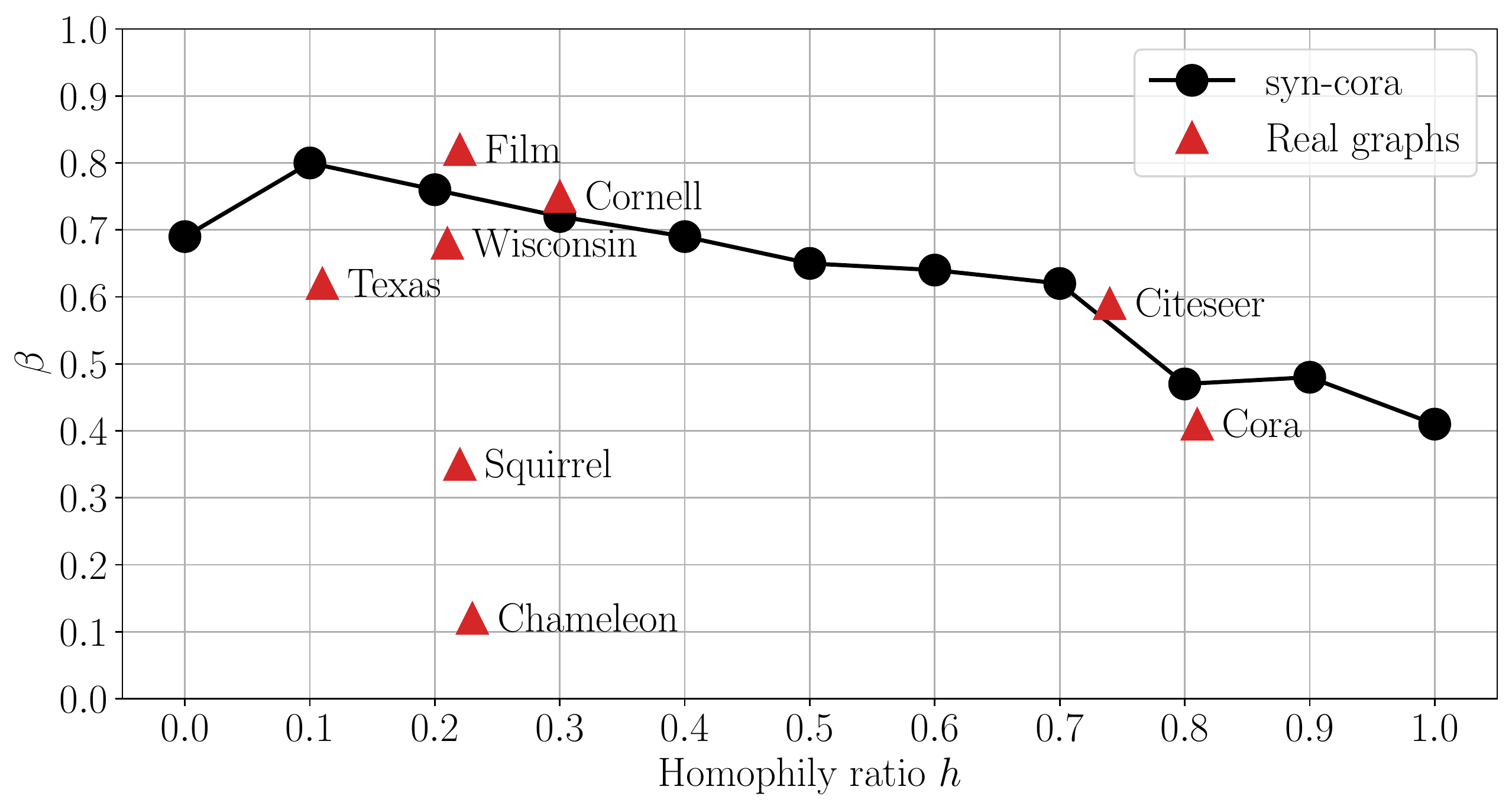}
\caption{Values of learned parameter $\beta$ for GCNH on real and synthetic datasets. Results are obtained using the best-performing hyperparameter configuration for GCNH models with one layer.}
\label{fig:gcnh_beta_analysis}
\end{figure}

\subsection{Analysis of the GCNH Design}\label{sec:experiments-analysis}
To measure the impact of the design choices we made for GCNH, we perform an ablation study by removing the two main components of GCNH, namely, the separate MLPs and the learned importance coefficient $\beta$, to see how they affect the performance.
We report the results in Table~\ref{tab:ablation_exp}, where we use a standard GCN as a baseline.
The table shows how both components bring an improvement in classification accuracy on heterophilous graphs, with a small tradeoff on the performance on homophilous graphs.
The MLPs separation leads to the largest gain, and its combination with the $\beta$ coefficient consistently brings further improvements.

We analyze further the behavior of $\beta$.
Since $\beta$ balances the contribution of messages of self-node and its neighbors, we expect heterophilous graphs to lead to larger $\beta$s given that neighborhoods are generally less informative in those cases.
Figure~\ref{fig:gcnh_beta_analysis} shows the values of the parameter $\beta$ on different graphs.
On synthetic graphs, $\beta$ is evidently correlated with the edge homophily ratio measured by $h$.
On most real graphs, the values follow a trend similar to that observed in synthetic graphs.
Chameleon and Squirrel are exceptions: we find that, while these datasets are heterophilous according to the $h$ metric, GCNH tends to learn low values of $\beta$, corresponding to highly informative neighborhoods.
This corroborates the findings outlined in~\cite{ma_is_2022}, which pointed out a similar contradiction: how a standard GCN performs unexpectedly well on the heterophilous Chameleon and Squirrel, even outperforming heterophily-specific methods.
In fact, this result further suggests that edge homophily ratio is not suited to describe neighborhood informativeness in general.

\subsection{Aggregation Functions}
\label{sec:agg_func}
We test three different aggregation functions for $\bigoplus_{v\in \mathcal{N}_u}$ in Equation~\eqref{eq:gcnh_feat_trans2b}: element-wise sum, element-wise mean across the neighbors and element-wise max. 
Table~\ref{tab:aggfunc_exp} reports a comparison of the node classification results for these aggregation functions.

\begin{table}
    \centering
    \footnotesize
    \addtolength{\leftskip} {-2cm}
    \addtolength{\rightskip}{-2cm}
    \setlength{\tabcolsep}{0.3em}
    \begin{tabular}{ c c | c c c c c}
    \toprule
     & \textbf{N. Params} & \textbf{Cora} & \textbf{Citeseer} & \textbf{Chameleon} & \textbf{Squirrel} & \textbf{Film} \\
    \midrule
    % GPRGNN & 16.87 & 22.11 & 19.57 & 30.70 & 18.02 \\
    \textbf{GGCN} & 118~k & 96.05 & 99.94 & 74.83 & 331.64 & 628.36  \\
    \textbf{O(d)-SD} & 46~k & 19.64 & 20.15 & 48.87 & 275.90 & 44.22 \\
    \midrule
    \textbf{GCNH} (sum) & 30~k & \textbf{8.79} & \textbf{10.28} & \textbf{8.56} & \textbf{12.61} & 15.59 \\
    % GCNH-max & 40.69 & 59.96 & 33.06 & 141.79 & 279.20 \\
    \textbf{GCNH} (max) & 30~k & 11.26 & 13.19 & 10.71 & 17.20 & \textbf{12.82}\\
    \bottomrule
    
    \end{tabular}
    \caption{Number of trainable parameters and training times (sec) for two state-of-the-art methods and GCNH on several graphs. Training is performed for 200 epochs on 10 splits for each dataset. Models have one layer and hidden size 16. Shortest times for each dataset are in \textbf{bold}.}
    \label{tab:training times}
\end{table}

% \begin{figure}
% \centering
% \includegraphics[width=.24\textwidth]{images/trainable_params.pdf}
% \caption{Number of trainable parameters for GGCN, O(d)-SD and GCNH; $y$-axis is on a log scale. All models have one layer and hidden size 16.}
% \label{fig:trainable_params}
% \end{figure}

\subsection{GCNH And Oversmoothing}
\label{sec:oversmoothing}
\begin{figure}
% \centering
\begin{subfigure}{.24\textwidth}

    \includegraphics[width=\textwidth]{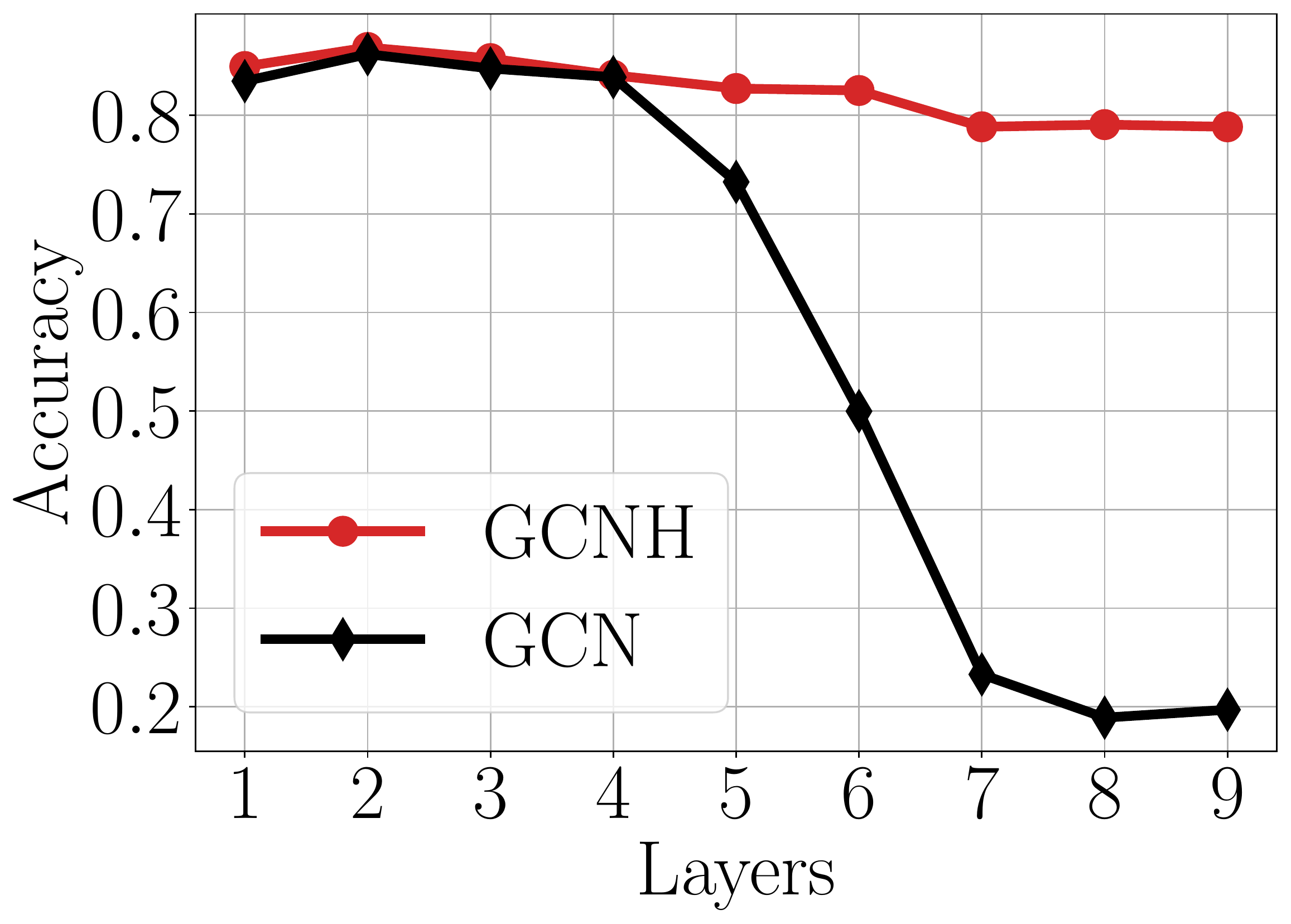}
    \caption{Cora}
    \label{fig:oversmoothing_cora}
    \end{subfigure}
\hfill
\begin{subfigure}{.24\textwidth}
    \includegraphics[width=\textwidth]{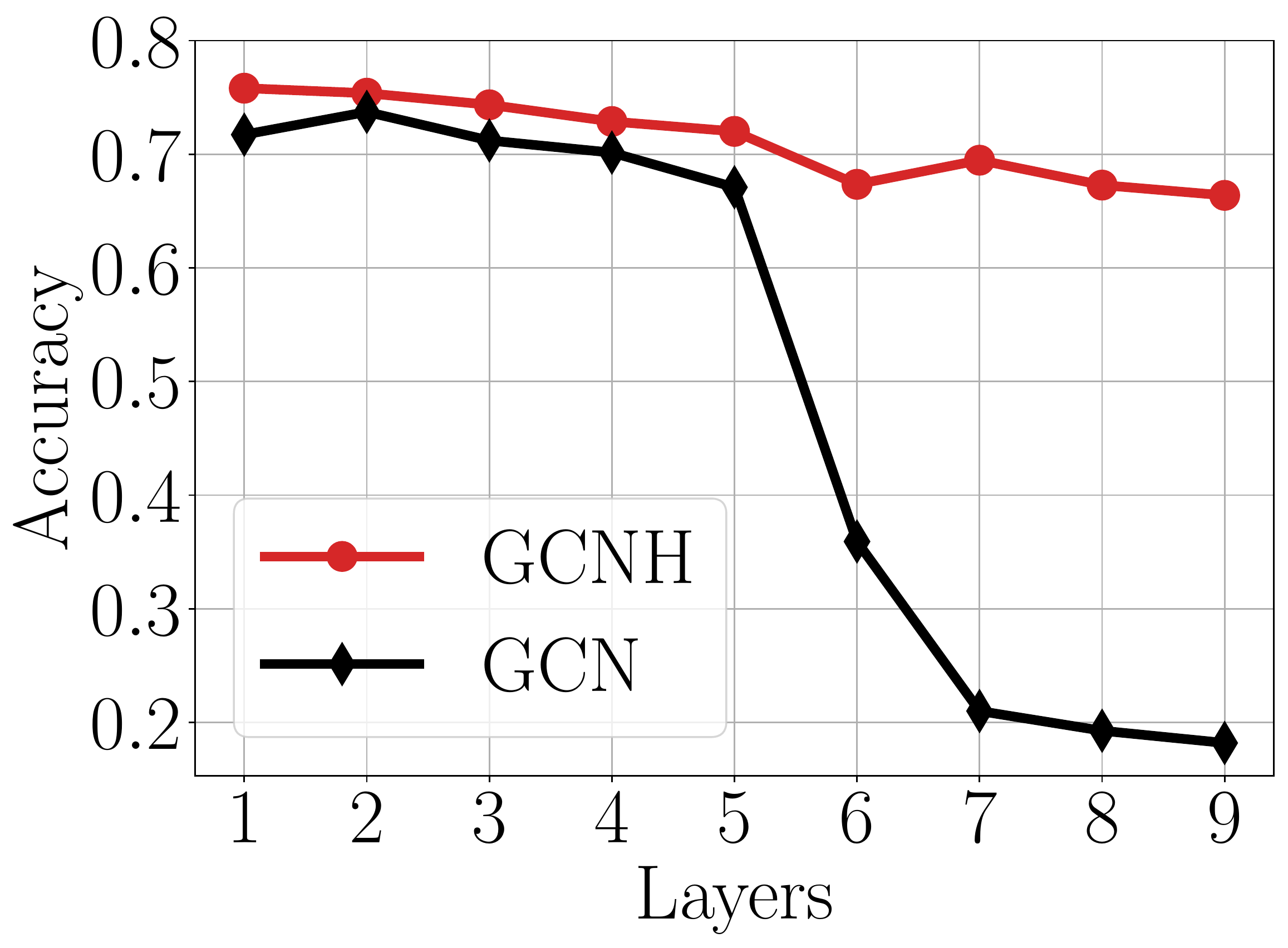}
    \caption{Citeseer}
    \label{fig:oversmoothing_citeseer}
\end{subfigure}
% \hfill

\caption{Accuracy for GCNH and GCN with different numbers of layers. With a large number of layers, GCNH prevents oversmoothing.}
\label{fig:oversmoothing}
\end{figure}

The performance of GNNs is known to gradually decrease when increasing the number of layers. This decay is partly attributed to oversmoothing, i.e., repeated graph convolutions eventually making node embeddings indistinguishable from each other~\cite{li_insights_2018, oono_graph_2020}.
We show experimentally that the design choices of GCNH  alleviate the oversmoothing problem. As shown in Figure~\ref{fig:oversmoothing}, GCNH's accuracy decreases just slightly when increasing the number of layers, whereas increasing the layers of GCN leads to a larger drop in performance; see Section~\ref{app:sec:oversmoothing} in Appendix for the experimental details.

\subsection{Training Times And Trainable Parameters}
\label{sec:training_times}
We report in Table~\ref{tab:training times} the training times required by GCNH and two of its main competitors GGCN~\cite{yan_two_2021} and O(d)-SD~\cite{bodnar_neural_2022}, as well as their number of trainable parameters. 
We take the implementations of these methods from the repositories of the authors~\cite{yujun-yan-code,neural-sheaf-diffusion-code}.
We report the result for GCNH using sum and max aggregation. We omit results for mean since sum and mean require the same amount of computation as they can be implemented efficiently with a matrix multiplication, i.e., their training times are the same. Max, instead, requires a scattered max pooling operation for which we use the implementation in~\cite{pytorch-scatter}.
GCNH is noticeably faster and has fewer trainable parameters than both GGCN and O(d)-SD.

\section{Conclusions}
We introduced GCNH, a simple yet effective GNN architecture that improves representation capabilities on heterophilous graphs. 
GCNH leverages two design choices: (i) two distinct learnable mapping functions that separately encode the center node and its neighbors into intermediate representations and (ii) the importance coefficients $\beta$, one per layer, to balance the contributions of these two representations on the final node embeddings.
We analyze and demonstrate how these two components enhance the representation capabilities of GCNH compared to GCN, as, together, they reduce the detrimental contributions of noisy neighborhoods to the updated node representations, while making effective use of basic structural information beneficial for classification. 
Through extensive experiments on real and synthetic graphs, we show that GCNH performs competitively on the node classification task, outperforming state-of-the-art methods on four out of eight common real-world datasets.  
In addition to its effectiveness, GCNH's simple design results in significantly faster training times and fewer trainable parameters compared to competing methods.

\appendices

\vspace{6cm}

\begin{table*}[t]
    \centering
    \footnotesize
    \addtolength{\leftskip} {-2cm}
    \addtolength{\rightskip}{-2cm}
    %\setcellgapes{3pt}
    %\makegapedcells
    \begin{tabular}{c c c c c c c c c c }
    \toprule
    \textbf{Benchmark} & \textbf{Cornell} & \textbf{Texas} & \textbf{Wisconsin} & \textbf{Film} & \textbf{Chameleon} & \textbf{Squirrel} & \textbf{Cora} & \textbf{Citeseer} & \textbf{syn-cora} \\
    \midrule
    \textbf{\#Nodes} & 183 & 183 & 251 & 7,600 & 2,277 & 5,201 & 2,708 & 3,327 & 1490 \\
    \textbf{\#Edges} & 280 & 295 & 466 & 26,752 & 31,421 & 198,493 & 1,433 & 3,703 & 2965 to 2968 \\
    \textbf{\#Classes} & 5 & 5 & 5 & 5 & 5 & 5 & 7 & 6 & 5 \\
    \textbf{\#Features} & 1,703 & 1,703 & 1,703 & 931 & 2,325 & 2,089 & 1,433 & 3,703 & 1,433 \\
    \textbf{Homophily \(h\)} & 0.30 & 0.11 & 0.21 & 0.22 & 0.23 & 0.22 & 0.81 & 0.74 & [0, 0.1, ..., 1] \\
    \bottomrule

    \end{tabular}
    \caption{Statistics of real-world and synthetic datasets.}
    \label{tab:data_stat}

\end{table*}

\begin{table*}[t]
    \centering
    \footnotesize
    \addtolength{\leftskip} {-1cm}
    \addtolength{\rightskip}{-1cm}
    \begin{tabular}{ c c c c c c c }
    \toprule
    \textbf{Dataset} & \textbf{\#Layers} & \textbf{Batch size} & \textbf{\#Epochs} & \textbf{Hidden size} & \textbf{Dropout rate} & $\beta$ \\
    \midrule
    \textbf{Cornell} & \{\textbf{1}, 2, 3\} & \{\textbf{50}, $|V|$\} & \{100, 200, \textbf{300}\} & \{\textbf{16}, 32, 64\} & \{0.0, \textbf{0.25}, 0.5\} & 0.75 \\
    \textbf{Texas} & \{\textbf{1}, 2, 3\} & \{50, $\boldsymbol{|V|}$\} & \{100, 200, \textbf{300}\} & \{16, \textbf{32}, 64\} & \{0.0, \textbf{0.25}, 0.5\} & 0.62 \\
    \textbf{Wisconsin} & \{1, \textbf{2}, 3\} & \{\textbf{50}, $|V|$\} & \{100, 200, \textbf{300}\} & \{16, \textbf{32}, 64\} & \{0.0, \textbf{0.3}, 0.6\} & 0.70, 0.69\\
    \textbf{Film} & \{1, \textbf{2}, 3\} & \{300, \textbf{500}, $|V|$\} & \{\textbf{150}\} & \{16, \textbf{32}, 64\} & \{0.0, 0.3, \textbf{0.6}\} & 0.72, 0.75 \\
    \textbf{Chameleon} & \{\textbf{1}, 2, 3\} & \{\textbf{300}, 600, $|V|$\} & \{500, \textbf{1000}, 1500\} & \{16, \textbf{32}, 64\} & \{\textbf{0.0}, 0.25, 0.5\} & 0.12 \\
    \textbf{Squirrel} & \{\textbf{1}, 2, 3\} & \{300, \textbf{1400}, $|V|$\} & \{500, 1000, \textbf{1500}\} & \{16, \textbf{32}, 64\} & \{\textbf{0.0}, 0.25, 0.5\} & 0.35 \\
    \textbf{Cora} & \{1, \textbf{2}, 3\} & \{\textbf{150}, 300, $|V|$\} & \{\textbf{300}\} & \{16, 32, \textbf{64}\} & \{0.0, 0.25, 0.5, \textbf{0.75}\} & 0.69, 0.61 \\
    \textbf{Citeseer} & \{\textbf{1}, 2, 3\} & \{150, \textbf{300}, $|V|$\} & \{\textbf{300}\} & \{\textbf{16}, 32, 64\} & \{0.0, \textbf{0.25}, 0.5, 0.75\} & 0.59 \\
    \textbf{syn-cora} & \{1, 2, 3\} & \{300, $|V|$\} & \{300\} & \{16, 32\} & \{0.0, 0.25, 0.5\} \\
    \bottomrule
    
    \end{tabular}
    \caption{Hyperparameters and learned values of $\beta$ for GCNH. For each dataset, the best hyperparameters are in \textbf{bold}. For 2-layer models, the values of $\beta$ are in the order $\beta^0, \beta^1$. For syn-cora, the best hyperparameters and the values of $\beta$ are not reported as they depend on the homophily value of the graph.} 
    \label{tab:gcnh_hyp}
\end{table*}

\newpage
\section{Datasets}
\label{app:datasets}

This section and Table~\ref{tab:data_stat} report details about the datasets.

\subsection{Real-World Datasets}
\label{sec:real_world_datasets}
These datasets are commonly used in most of the works dealing with heterophilous graphs.
Chameleon and Squirrel are taken from~\cite{simplifying-code}, the others from~\cite{geom-gcn-code}.
We present further details in the following:
\begin{itemize}[leftmargin=.5cm]
    \item \textbf{Texas, Wisconsin and Cornell} are webpage datasets collected from different universities~\cite{webkbproject}. Nodes represent web pages and edges are hyperlinks between them. %Node features are bag-of-words representations of the web pages, which are manually classified into five categories: student, project, course, staff, faculty. 
    \item \textbf{Film} (\textbf{Actor}), is the actor-only induced subgraph of the film-director-actor-writer network \cite{tang_social_2009}. Nodes correspond to Wikipedia pages of actors and edges denote the co-occurrence of two actors on the same page. %Node features are some keywords in the Wikipedia pages and labels are assigned by \cite{pei_geom-gcn_2019} based on words of the actors' Wikipedia pages.
    \item \textbf{Chameleon and Squirrel} are Wikipedia pages on the specific topics of chameleons and squirrels. %They were collected by \cite{rozemberczki_multi-scale_2021} and pre-processed by \cite{pei_geom-gcn_2019}. 
    Nodes are Wikipedia pages and edges are mutual links between them. 
    %Node features indicate the presence of informative nouns on Wikipedia pages. Nodes are classified into five categories based on the average monthly traffic on the web page.
    \textit{Note:} contemporary to this work,~\cite{platonov2023critical} highlights drawbacks of these two datasets.
    \item \textbf{Cora and Citeseer} are citation networks where nodes represent papers and edges represent citations of one paper by another \cite{sen_collective_2008, namata_query_2012}. %Node features are bag-of-words representations of papers and labels are the academic topics of the papers.
    These datasets are treated as undirected.
\end{itemize}

\subsection{Synthetic Datasets}
\label{app:dataset:syn}
The synthetic datasets used in this work are taken from \cite{zhu_beyond_2020}, which defines a graph generation strategy similar to \cite{abu_mixhop_2019}. The homophily ratio of the graph is defined \textit{a priori}. Then, the graph is generated such that the degree distribution follows a power law. Classes are assigned randomly and features are sampled from nodes of the corresponding class in the real graph Cora. For each of the specified homophily levels, three different graphs are generated.

\section{Hyperparameters And Experimental Details}
\label{app:hyp}
We report here additional details about the experiments in this work. All models are trained with Adam optimizer, learning rate $5\cdot10^{-3}$ and weight decay $5\cdot10^{-3}$. 
\subsection{Baselines On Synthetic Datasets}
\label{app:baselines_syn_hyp}
We obtain the results for MLP, GCN and GAT on the synthetic graphs in Figure \ref{fig:gcnh_syn} through hyperparameter optimization among the following values: 100 epochs, \{1,2,3\} layers, hidden size \{16,32\}. For all experiments on the synthetic datasets, we randomly generate train/evaluation/test splits with sizes 50\%/20\%/30\%. We compute results on a single split for each dataset and we average the results of three different graphs for each level of homophily ratio reported in the figure.
\subsection{GCNH}
\label{app:gcnh_hyp}
The hyperparameters for GCNH on real and synthetic graphs are optimized among the values reported in Table \ref{tab:gcnh_hyp}. The activation function $\sigma(\cdot)$ in Equations \ref{eq:gcnh_feat_trans1b} and \ref{eq:gcnh_feat_trans2b} is LeakyReLU. We perform batching by forwarding, for each batch, the complete graph through the model and computing the loss only on the nodes in the current batch.

\subsection{Experiments On Oversmoothing}\label{app:sec:oversmoothing}
The hyperparameters of the models in Section~\ref{sec:oversmoothing} are optimized among the following values: batch size 300, epochs \{300,500\}, hidden size \{16,32,64\}, dropout rate \{0.0,0.5\}. Note that the results for GCN in Table \ref{tab:main_results} are taken from~\cite{yan_two_2021}, while those in Figure \ref{fig:oversmoothing} are obtained from our own experiments, and they slightly differ because the hyperparameter grid search is different.

\end{document}